# Flight Demand Forecasting with Transformers

Liya Wang,[1] Amy Mykityshyn,[2] Craig Johnson,[3]
*The MITRE Corporation, McLean, VA, 22102, United States*

Jillian Cheng[4]
*Federal Aviation Administration*

**Transformers have become the de-facto standard in the natural language processing (NLP) field. They have also gained momentum in computer vision and other domains. Transformers can enable artificial intelligence (AI) models to dynamically focus on certain parts of their input and thus reason more effectively. Inspired by the success of transformers, we adopted this technique to predict strategic flight departure demand in multiple horizons. This work was conducted in support of a MITRE-developed mobile application, Pacer, which displays predicted departure demand to general aviation (GA) flight operators so they can have better situation awareness of the potential for departure delays during busy periods. Field demonstrations involving Pacer's previously designed rule-based prediction method showed that the prediction accuracy of departure demand still has room for improvement. This research strives to improve prediction accuracy from two key aspects: better data sources and robust forecasting algorithms. We leveraged two data sources, Aviation System Performance Metrics (ASPM) and System Wide Information Management (SWIM), as our input. We then trained forecasting models with temporal fusion transformer (TFT) for five different airports. Case studies show that TFTs can perform better than traditional forecasting methods by large margins, and they can result in better prediction across diverse airports and with better interpretability.**

## I. Nomenclature

$T$ = time interval between two consecutive time steps, 15 minutes
$t$ = time step
$p$ = number of time lags
$\tau$ = steps from current time t
$\tau_{max}$ = maximum look ahead times
$X_t$ = observed inputs at time t
$F_t$ = future known inputs at time t
$y_t$ = response variable at time t
$mae$ = mean absolute error
$mse$ = mean squared error

[1] Artificial Intelligence, Lead, Department of Operation Performance
[2] Aviation Systems Engineering, Lead, Department of Safety Intelligence Concepts and Evolution
[3] Group Leader, Department of NAS Future Vision & Research
[4] Project Manager, Surveillance Branch, ANG-C51

## II. Introduction

Airport surface congestion and flight delays due to sudden surges in General Aviation (GA) flight demand have been a concern for airlines, airport operators, and the Federal Aviation Administration (FAA). Surface demand is comprised of aircraft that operate as scheduled commercial flights or unscheduled GA operations. Commercial flights are scheduled months in advance and their schedule is shared among the airlines, the airports, and the FAA, whereas GA flights schedules are often more flexible, and therefore, less predictable. For airports that experience large increases in GA traffic due to special public events (e.g., professional sports games) or airports that have a consistently high percentage of GA traffic (e.g., Van Nuys Airport (VNY), Teterboro Airport (TEB)), it can be quite challenging for FAA traffic managers to strategically manage expected demand.

The MITRE Corporation's Center for Advanced Aviation System Development (MITRE CAASD) has developed Pacer [1], a research prototype mobile application, to support the FAA in exploring methods for efficient surface management. Pacer displays future departure demand information to GA pilots and allows pilots to provide updates to their intended departure times; this information is subsequently used to improve future departure demand predictions. Field demonstrations involving Pacer's rule-based prediction method showed that the prediction accuracy of departure demand still has room for improvement. Therefore, this research is one of several efforts that are exploring ways to improve prediction accuracy for GA departure demand. This paper demonstrates our efforts to improve Pacer's demand prediction by utilizing the cutting-edge Transformers technique.

Deep learning (DL) models have grown in popularity over traditional forecasting methods due to characteristics such as flexible learning ability, versatility, handling heterogeneous inputs, and having attention mechanisms to capture complex time dependencies between inputs and outputs (e.g., [2], [3], [4], [5], [6], [7], [8], [9], [10]). Since the inception of Transformers in 2017 [11], Transformers have become the de-facto standard in the natural language processing (NLP) field. They also started being used in computer vision [12] and other domains. Transformers can enable Artificial Intelligence (AI) models to dynamically focus on certain parts of their input and thus reason more effectively.

Aviation flight demand prediction is a real-world application of a multi-horizon multi-variable time series forecasting problem, which are challenging due to multiple heterogeneous input variables, including observed past ground information and future known information such as hour of day, day of week, and month, etc. To tackle these complicated problems, our research explores the use of a temporal fusion transformer (TFT) [5], which is an innovative attention-based architecture, which combines high-performance multi-horizon forecasting with interpretable insights into temporal dynamics. In addition, TFT has been proven to achieve significant performance improvements over a variety of established time series forecasting problems [2], [5].

The remainder of this paper is organized as follows: Section III gives a short introduction about our input data sources used in the research and Section IV describes our problem formulations. Section V provides details of the deep learning modeling process, and the results are shown in Section VI. Finally, we summarize our conclusions in Section VII.

## III. Data Source

DL algorithms run on data and the need for large amounts data is greater than ever. But more than large amounts of data, good data quality is also crucial to get the desired end result. Therefore, we took great effort to prepare our modeling dataset.

For our departure demand prediction analysis, we leveraged two aviation datasets: 1) ASPM [13] and 2) SWIM [14]. Short descriptions of these two data sources are below.

### A. Aviation System Performance Metrics (ASPM) Dataset

The ASPM dataset is a well-known operational data source that the FAA maintains. It integrates multiple data sources and provides information about flights arriving or departing an airport. The ASPM dataset includes flight counts, scheduled flight demand, delays, runway configuration, and weather information aggregated into 15 minutes bins. Table 1 lists the key data items used in our study [13]. Specifically, quarter-hour departure demand (DEPDemand) is our targeted forecasting objective, and the other elements are our inputs. One shortcoming with the ASPM dataset is its long updating frequency, only once per day. In addition, only the previous day's data are available. Therefore, ASPM data is not well suited for frequent forecasting updates.

Table 1 Data Items from ASPM

| Column Name | Description |
| --- | --- |
| Slice_Start_UTC | 15-minute period start time in UTC time |
| Hour | UTC hour (0-23), derived from Slice_Start_UTC |
| Qtr | UTC quarter hour (1 to 4), derived from Slice_Start_UTC |
| Day of week | UTC day of week (1 to 7), derived from Slice_Start_UTC |
| Month | UTC month (1 to 12), derived from Slice_Start_UTC |
| DEPDemand | Number of aircraft intending to depart for the period |

B. System Wide Information Management (SWIM) Dataset

To complement the ASPM dataset's deficiencies, we incorporated another FAA data source, SWIM, which provides first-hand, near real-time information about the state of an airport's surface. SWIM data are updated every minute and include information such as arrival/departure lists and taxi-in/taxi-out times. The departure list includes surface track information such as latitude, longitude, and time for active flights which we used to track observed departure flight counts. With this dataset, our modeler can reduce prediction error by 1) using shorter look ahead times, and 2) updating the forecasts more frequently.

SWIM data is provided in Extensible Markup Language (XML) format, and then each user group processes it according to their research needs. The version of SWIM data available to support our work is noisy. Therefore, we cleaned the data through departure track assembly, data exploration, data visualization, and domain experts' evaluation. Fig. 1 demonstrates our data cleaning process.

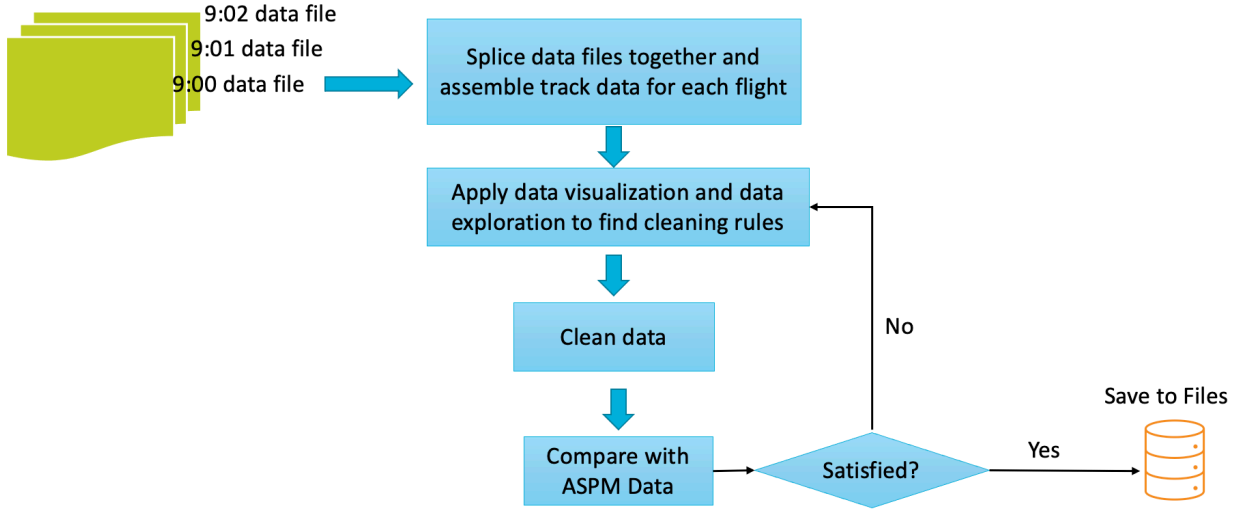

Fig. 1 SWIM data cleaning process.

In addition, the version of SWIM data we used covers only the Core 30 airports [15]. Therefore, we built two types of models for airports: one that can use both ASPM and SWIM data as inputs and another that just uses ASPM data when SWIM data are not available.

IV. Problem Formulation

To predict departure flight demand, we gathered the data into 15-minute bins accounting for future objectives and past information. Specifically, data included the objectives' own historical information, exogeneous time series data, such as observed surface departure flight counts, and known factors such as hour of day, and day of the week. The problem can be formulated as multivariate multi-step time series forecasting problems (Eq. (1)). Time lag ($p$) was learned from the data, and the maximum look ahead time, $\tau_{max}$, was determined by the data updating frequency and Pacer's needs. Fig. 2 explicitly demonstrates the relationship of inputs and outputs across a time horizon. The inputs are made up of two parts: past observed inputs, which includes past targets, and future known inputs. For our targeted departure demand (DEPDemand) prediction, Table 2 lists the inputs and outputs. Next, we will discuss how deep learning methods, TFT as well others, are built to find function $f$.

$$y_{t+\tau} = f(y_{t+\tau-1}, y_{t+\tau-2}, \ldots, y_{t-p}, X_{t-1}, X_{t-2}, \ldots, X_{t-p}, F_{t+1}, F_{t+2}, \ldots, F_{t+\tau}) \quad (1)$$

where $\tau = [1, \ldots, \tau_{max}]$

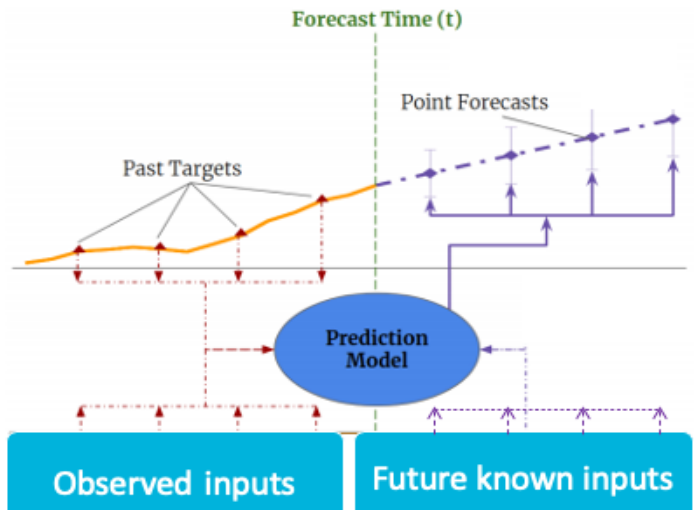

**Fig. 2 Illustration of forecasting with observed inputs and future known inputs [5].**

**Table 2 Inputs and outputs for departure demand prediction**

| **Target output** | • DEPDemand (from ASPM) |
|---|---|
| **Observed inputs** | • Observed number of departure flights (from SWIM) <br> • y's history (from ASPM) |
| **Future known inputs** | • Hour of day (from ASPM) <br> • Quarter of hour (from ASPM) <br> • Day of week (from ASPM) <br> • Month (from ASPM) |

## V. Modeling Methods

Solving real-world time series forecasting problems is challenging. Aviation flight departure demand prediction is no exception. The challenges are multifaceted and include heterogeneous input variables, accounting for future known factors (e.g., day of week), difficult feature selection (e.g., whether arrival demand should be included as an input feature), and the unknown interactions between inputs. This research addresses the complicated time series forecasting problems with TFTs, selected for their superior capabilities. For example, they can handle multiple exogenous variables with complex dependencies, identify the multifaceted relationship between input variables and output variables, learn and adapt, and extend to multiple sites with ease. Fig. 3 depicts the designed DL model training architecture for Pacer's demand forecasting. The following sections will present important details for building forecasting models.

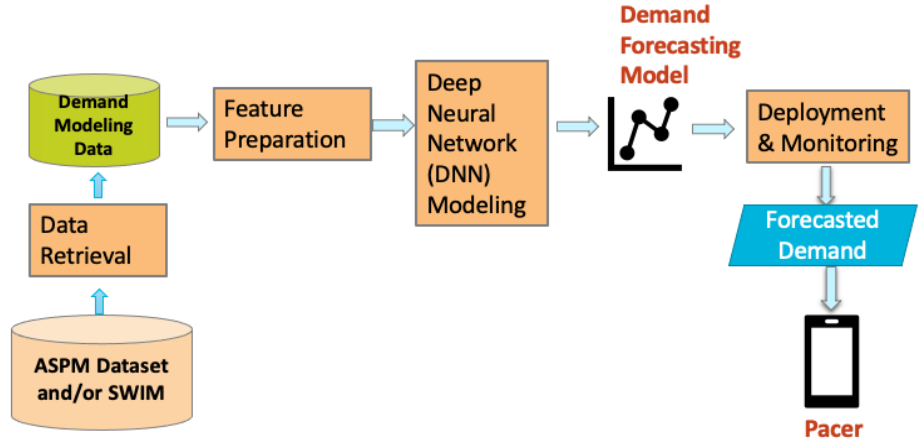

Fig. 3 Deep learning architecture for Pacer's demand and delay forecasting.

A.  **Feature Preparation**

Feature preparation plays an essential role in training DL models. Simply stated, feature preparation transforms raw data into the DL algorithm-required format, and it also improves the model performance. Domain knowledge and feature engineering knowledge are key at this step. We designed a five-stage feature processing procedure (Fig. 4) to transform raw time series data into sequence data required by our sequence to sequence (seq2seq) and seq2seq with attention models, which are described in Sections D and E, respectively.

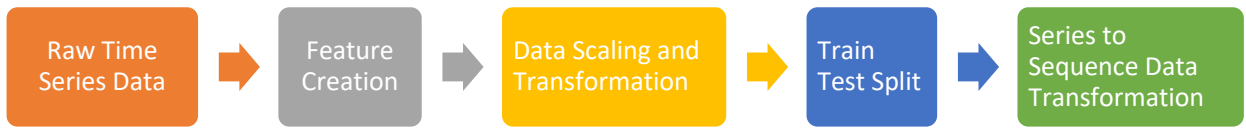

Fig. 4 Feature processing procedure for Pacer demand forecasting.

The important feature processing steps are explained as follows:
- **Feature Creation**: It is a process of identifying and transforming relevant data attributes into useful input. Analysts often employ feature creation to increase the value of available data. In this step, we used a data column from ASPM (SLICE_START_UTC) to create hour of day, day of week, quarter hour, and month categorical features with consideration of those factors that have an impact on demand.
- **Data Scaling and Transformation**:  Scaling is an important step towards making inputs unitless and in the same range. The most common techniques of feature scaling are normalization and standardization. We used a min-max scaler to transfer numerical features into a range of [0,1]. Another important characteristic of feature scaling is that neural network gradient descent converges much faster with feature scaling than without it. Fig. 5 compares the gradient descent process with and without scaling. The parameters are defined as follows: $x_1$ and $x_2$ are input variables; $w_1$ and $w_2$ are weights to optimize; the blue contour represents the loss function error surface; and the red line is the optimization process. Without scaling, the optimization process has a zig-zag shape, which takes more steps to reach the minimum point than the one with scaling. For our categorical variables, we used an embedding layer to find their data representations (see Fig. 8 for illustration).

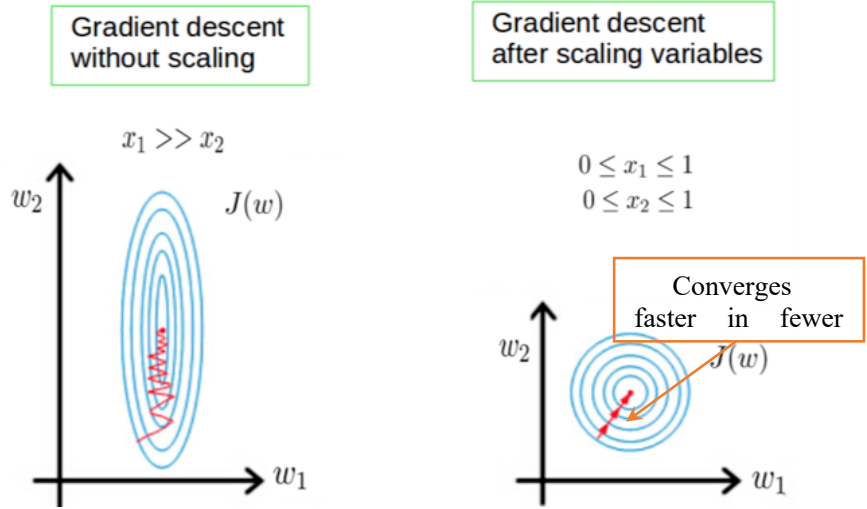

**Fig. 5 Illustration the importance of feature scaling [16].**

- **Train Test Data Split:** In this step, the dataset is usually split into a training dataset and a testing dataset. The model is first trained on the training dataset. Then the testing dataset is used to evaluate the model's performance and provides feedback on whether the model is overfitted or underfitted. A perfect model should be neither overfitted nor underfitted.

- **Time Series Data to Sequence Data:** The seq2seq with attention algorithm requires past sequence data X and future sequence data Y to construct a labeled dataset. For that, we adopted the common practice of applying a fixed length sliding time window approach (Fig. 6**Error! Reference source not found.**) to construct X, Y pairs for model training and testing. We ran a sliding window on the transformed training set to get training samples with features and labels, which are the previous $p$ and next $\tau_{max}$ observations respectively. Test datasets are also constructed in the same manner for model evaluation.

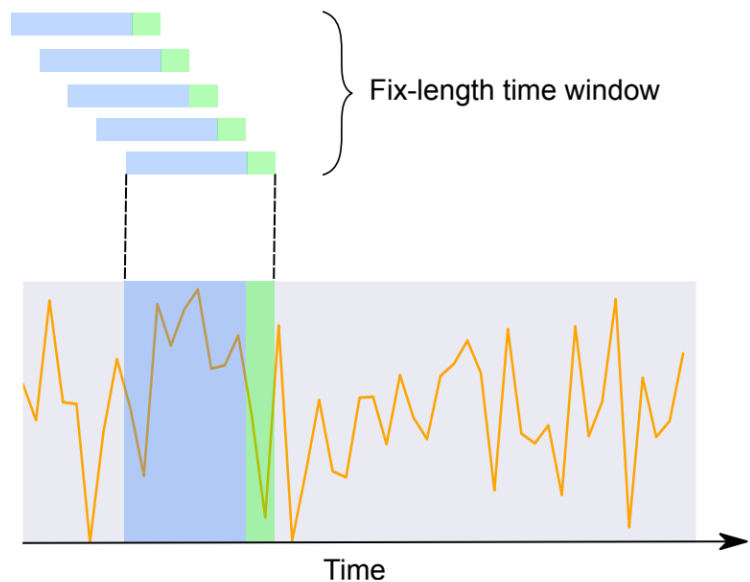

**Fig. 6 Construct supervised learning sequence data from time series data [4].**

To have a better understanding on the performance of TFT, we also ran four other algorithms: 1) Linear Regression (LR); 2) Autoregression (AR); 3) sequence-to-sequence (seq2seq); and 4) seq2seq with attention for the comparison. The following sections will have a short introduction for these five algorithms.

## B. Linear Regression (LR)

In statistics, linear regression is used to model the relationship between a scalar response and one or more explanatory variables. The very simplest case of a single scalar predictor variable *x* and a single scalar response variable *y* is known as simple linear regression (see Fig. 7). The extension to multiple and/or vector-valued predictor variables (denoted with *X*) is known as multiple linear regression, also known as multivariable linear regression. Nearly all real-world regression models are multiple regression models. Note, however, that in these cases the response variable *y* is still a scalar. Another term, multivariate linear regression, refers to cases where *y* is a vector [17].

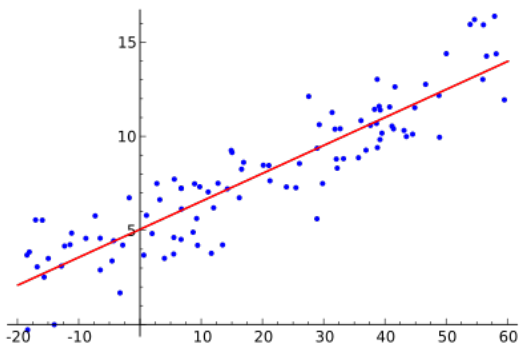

**Fig. 7 Example of simple linear regression, which has one independent variable [17].**

### C. Autoregression (AR)

Autoregression (AR) is another popular stochastic process model designed for time series forecasting [18]. The autoregressive model specifies that the output variable depends linearly on its own previous values and on a stochastic term, as shown in Eq. (2). AR methods cannot consider as many factor variables as DL and LR do. The only prior knowledge required is its history.

$$y_t = c + \phi_1 y_{t-1} + \phi_2 y_{t-2} + \cdots + \phi_p y_{t-p} + \epsilon_t \qquad (2)$$

Where y is the observation, c is constant (intercepts), $\phi$ is coefficient, and $\epsilon$ is error item.

### D. Sequence to sequence (seq2seq)

Seq2seq was originally designed by Google for machine translation problems and has achieved a lot of success in tasks like machine translation. One famous example is Google Translate, which started using such a model in production in late 2016. These models can be found in two pioneering papers [19], [20]. Inspired by the success of seq2seq in the NLP field, researchers have explored seq2seq for the time series forecasting problems because they have a similar problem structure (e.g., [21]). Seq2seq can achieve better performance than traditional statistical time series forecasting methods, which usually require strong restrictions and assumptions.

As the name suggests, seq2seq takes an input sequence (e.g., sentences in English) and generates an output sequence (e.g., the same sentences translated to Chinese). It does so by using a recurrent neural network (RNN). Two commonly used types of RNN are Long Short-Term Memory (LSTM) and Gated Recurrent Units (GRU) because LSTM and GRU can better handle the vanishing gradient problem [22], which can cause the neural network to not be trained properly.

Seq2seq is made up of two parts: the *encoder* and the *decoder*. It is sometimes referred to as the **Encoder-Decoder Network** (Fig. 8). The two components are explained as follows:

- **Encoder:** Uses deep neural network layers and converts the input sequence to a corresponding hidden vector as an initial state to the first recurrent layer of the decoder part.

- **Decoder:** Takes the input from a) the hidden vector generated by the encoder, b) its own hidden states, c) previous step output, and d) future known factors *F* to produce the next hidden vector and finally predict the next output.

In our research, our future known inputs are categorical variables such as hour of day, day of week, month, etc. Therefore, we used an embedding layer to encode them first and then send them to LSTM cells. Because LSTM type of cells performs better than GRU, we chose to proceed with LSTM in our seq2seq model building.

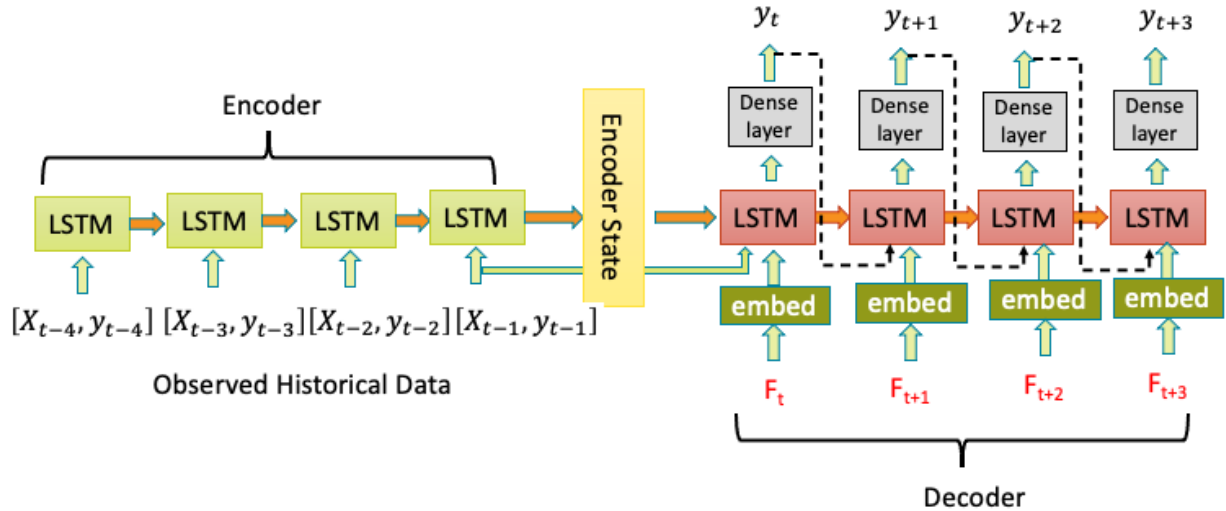

Fig. 8 Seq2seq architecture for our departure demand time series forecasting.

One of the drawbacks of seq2seq models is that they have difficulty handling long input sequences because their encoding context is wrapped up in a specified size vector. In the case of long sequences, there is a high probability that the initial context has been lost by the end of the sequence.

### E. Sequence to sequence (seq2seq) with Attention

To deal with the issue, seq2seq with attention models have been introduced in two papers ([23], [24]). As the name suggests, the method also introduced a technique called "Attention," which flexibly focuses on different parts of the input sequence at every stage of the output sequence by preserving all the context.

In psychology, attention is the cognitive process of selectively concentrating on one or a few things while ignoring others. The attention mechanism in neural networks is also an attempt to implement a similar action of selectively concentrating on a few relevant things, while ignoring others. Attention is one component of a network's architecture, and it manages and quantifies the interdependence between the input and output elements. The seq2seq model is unable to accurately process long input sequences because only the last hidden state of the encoder is kept as the context vector for the decoder [25]. The attention mechanism was proposed to directly address this issue and keeps all the hidden states of the input sequence. The attention mechanism creates a unique mapping between each time step of the decoder output and all the encoder hidden states. Therefore, the mechanism allows the model to focus and place more "attention" on the relevant parts of the input sequence as needed. Fig. 9**Error! Reference source not found.** illustrates our customized seq2seq with Luong's attention network structure (Fig. 10**Error! Reference source not found.**) for departure demand forecasting.

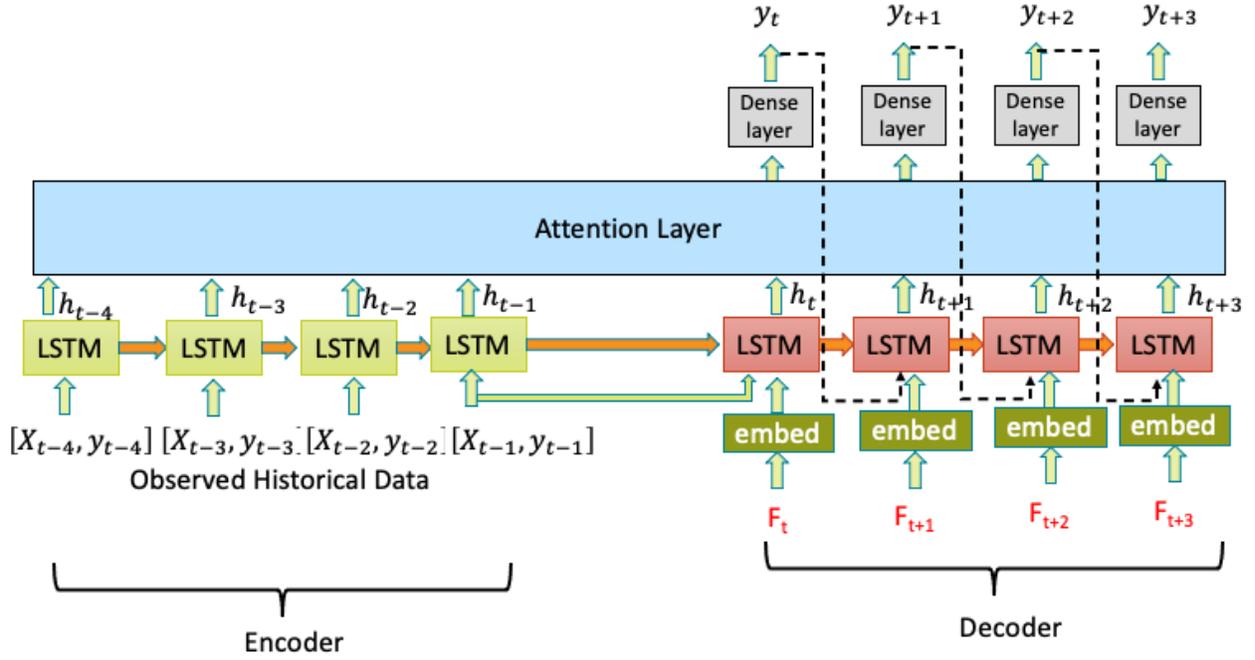

**Fig. 9 Seq2seq with attention architecture for our departure demand time series forecasting.**

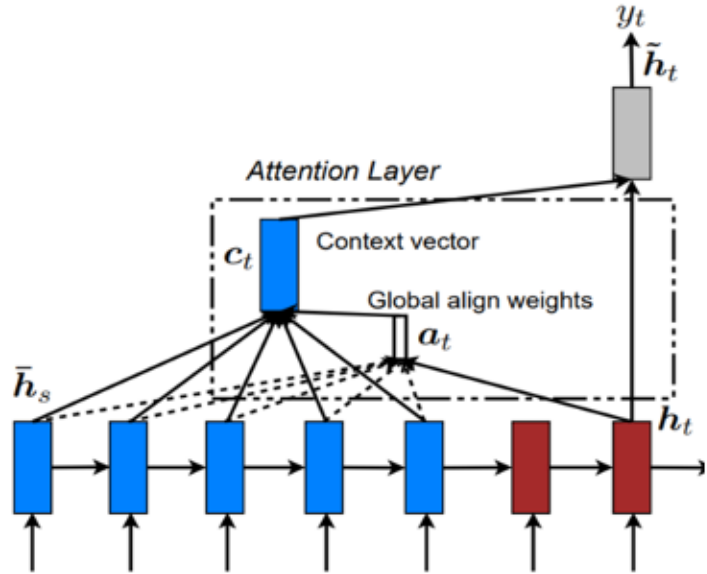

**Fig. 10 Luong's attention mechanism [23].**

### F. Temporal Fusion Transformer (TFT)

Similar to recurrent neural networks (RNNs), transformers can handle sequences of data like time series data. In addition, transformers have a position encoding mechanism which can free the requirement that the sequence data should be input in order. Due to this feature, the transformers can be trained in parallel [26], which can reduce training times greatly.

Encouraged by their versatility and superior performance, we employed Temporal Fusion Transformer (TFT) for our strategic departure demand prediction. TFT is an attention-based neural network structure specially designed to combine high-performance multi-horizon forecasting with explainable insights into temporal dynamics.

TFT's powerful submodules can efficiently learn data representations for each input type (i.e., static, known, observed inputs) and at the same time achieve high forecasting performance over a wide range of problems. Fig. 11 shows the architecture of TFT, and the major sub-components are explained as follows:

- **Gating mechanisms**: filter out any unused components of the architecture with adaptive depth and network complexity to accommodate a wide range of datasets and scenarios.
- **Variable selection networks**: select relevant input variables at each time step.
- **Static covariate encoders**: integrate static features into the network, through encoding of context vectors to condition temporal dynamics.
- **Temporal processing**: learn both long- and short-term temporal relationships from both observed and known time-varying inputs. A seq2seq layer is used for local processing, whereas a novel interpretable multi-head attention block is used to capture long-term dependencies.
- **Prediction intervals**: output quantile forecasts to determine the range of likely target values at each time step.

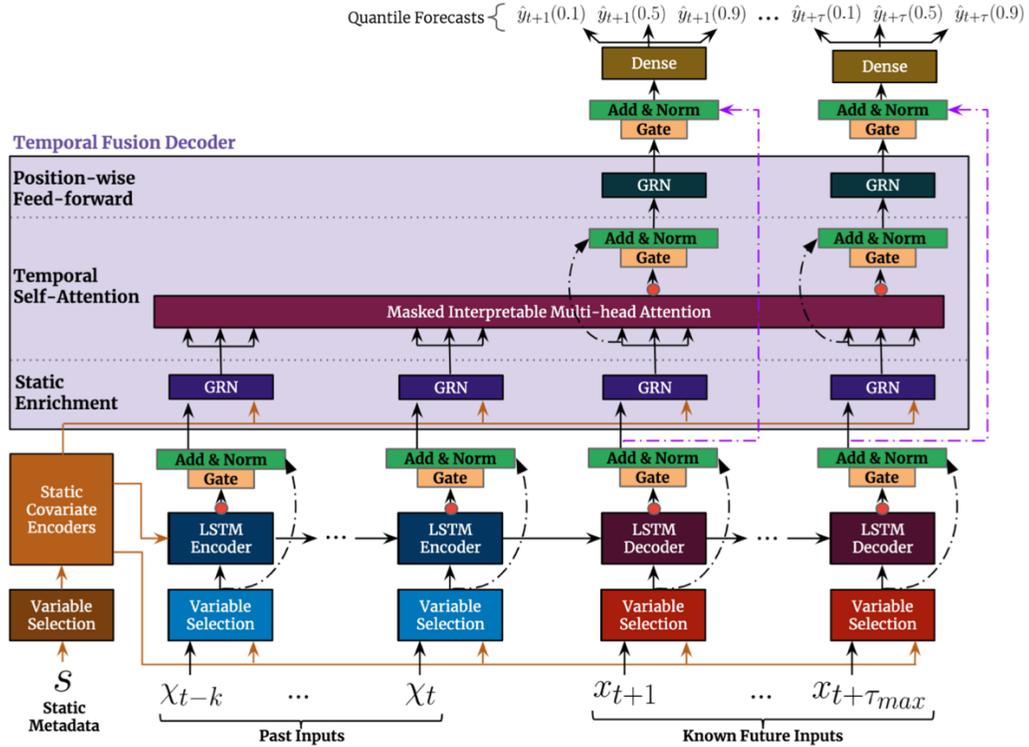

**Fig. 11 TFT Architecture [5].**

After presenting five modeling methods, next the modeling results will be presented and discussed.

## VI.  Results

This section presents a series of case study results. First, we selected Las Vegas McCarran International Airport (LAS) to test a set of forecasting techniques, comparing performance across six candidate models: linear regression (LR) and autoregression (AR), seq2seq, seq2seq with attention, TFT, and TFT using SWIM and ASPM input data sources. Based on the findings from the performance comparison, we tested the application of the best performing model on several other airports; namely, Charlotte Douglas International Airport (CLT), Dallas Love Field Airport (DAL), Van Nuys Airport (VNY), and Teterboro Airport (TEB), which were of interest to us given that each has a sizeable number of GA flight operations.

These airports were selected to assess Pacer under a variety of operational conditions. LAS represents a busy airport with many airline operations as well as a significant number of GA operations that can surge around certain events. CLT and DAL are two airports where we have deployed Pacer in partnership with NASA's Airspace Technology Demonstration 2 (ATD-2) project. Pacer is being used to allow GA flight operators to submit updated departure intent information, which is provided to NASA and used by the ATD-2 system to update its surface demand predictions. Since it can be difficult to get a consistent level of Pacer user participation, understanding how machine learning could be applied to those airports to improve departure demand predictions was of interest. VNY and TEB are two predominately GA airports that operate in the vicinity of large metroplex environments, Los Angeles, and New York City, respectively. The following sub-sections present a series of prediction results across these diverse airports.

### A. Train and Test Data Split
In ML, the dataset is commonly split into two sets: training and testing. For our model training, we used the calendar year of 2019 as our training dataset and one month of January 2020 as our testing dataset.

### B. LAS Departure Demand Forecasting Results
#### 1. LAS Forecasting Results
Table 3 compares the six models' performances with different combinations of data sources and modeling methods. The best evaluation metrics (mean squared error (mse) calculated in Eq. (3), mean absolute error (mae) formulated in Eq. (4), and explained variance) are highlighted in yellow. Mse and mae are commonly used evaluation metrics in machine learning regression problems to assess the model performance. For mse and mae values, a lower value is better and for the explained variance score, a higher value is better. Mae represents the average of the absolute value of the difference between the number of aircraft predicted by the model and the actual number of departed aircraft. Mse represents the average of the squared value of the difference in the number of predicted aircraft versus actual. In both metrics, the magnitude of the error is measured (i.e., the number of estimated departure flights when compared to truth), though the direction of the error (above or below the correct value) is not. The variables n_lag and n_look_ahead represent the number of quarter hours we look back and look forward respectively.

$$mse = \frac{1}{n}\sum_{i=1}^{n}(y_i - \hat{y}_i)^2 \qquad (3)$$

$$mae = \frac{1}{n}\sum_{i=1}^{n}|y_i - \hat{y}_i| \qquad (4)$$

where
- $n$ = number of data points
- $y_i$ = true value for data sample $i$
- $\hat{y}_i$ = predicted value for data sample $i$

According to Table 3, the TFT model, combined with input data sources of ASPM and SWIM, achieves the best performance. Its mse is 53% better than the AR model, and 31% better than the LR model. Even when a single data source (ASPM) is used, TFT still produces the predictions with the lowest error. In addition, TFT's performance with two input data sources does not differ much from the single data input source case (3% difference in mse, and equal mae), which indicates that the superiority of TFT can overcome some data deficiency. Determining if these errors are sufficiently small enough may depend on findings from future field evaluations.

Table 3 Model performances comparison

| data | model | mse | mae | explained_variance | n_lag | n_look_ahead | mse_comparison |
|---|---|---|---|---|---|---|---|
| ASPM | Linear_Regression | 7.63 | 2.1 | 0.65 | 10 | 124 | -31% |
| ASPM | Autoregressive | 8.91 | 2.3 | 0.59 | 96 | 124 | -53% |
| ASPM | Seq2Seq | 6.53 | 1.8 | 0.72 | 10 | 124 | -12% |
| ASPM | Seq2Seq_Attention | 6.27 | 1.8 | 0.72 | 10 | 124 | -7% |
| ASPM | TFT | 5.99 | 1.8 | 0.73 | 6 | 124 | -3% |
| ASPM&SWIM | TFT | 5.83 | 1.8 | 0.75 | 6 | 16 | 0% |

Elaborating further on some technical details of this analysis, it should first be noted that TFT can output multiple quantiles (e.g., 25th percentile, 50th percentile, 75th percentile) at each time step for the targeted objectives. We chose the 50th percentile as the predicted value to compare with other forecasting methods. Since SWIM data are updated every minute, we can shorten the look ahead times from 31 hours (124 quarter-hours) to 4 hours (16 quarter-hours) and at the same time update our predictions every 15 minutes, as demonstrated in Fig. 12. At UTC time 20:30, we predict demand for a 4-hour look ahead window of 20:30-00:30; at 20:45 (15 minutes later), we update the predicted demand with a new 4-hour look ahead window of 20:45-00:45. The graph shows that as time goes by, the updated prediction (green line) more closely represents the actual demand (red line), as compared to the initial prediction (blue line).

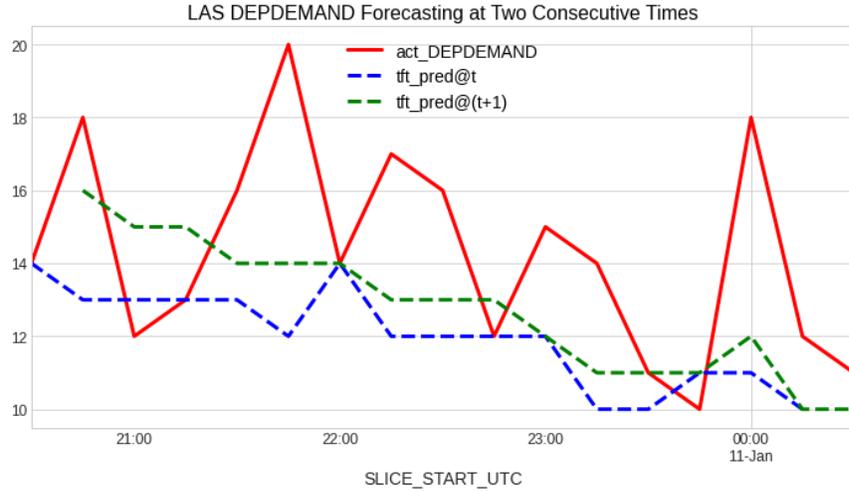

**Fig. 12 TFT predictions at two consecutive times.**

Fig. 13 shows an example of LAS quarter-hour departure demand forecasting results by the TFT model with both ASPM and SWIM as input data sources over several testing days. The red line represents the true departure demand, and the blue line is the forecasted demand by the TFT model. From the graph, we can see that the TFT model captures the up/down trends of demand well.

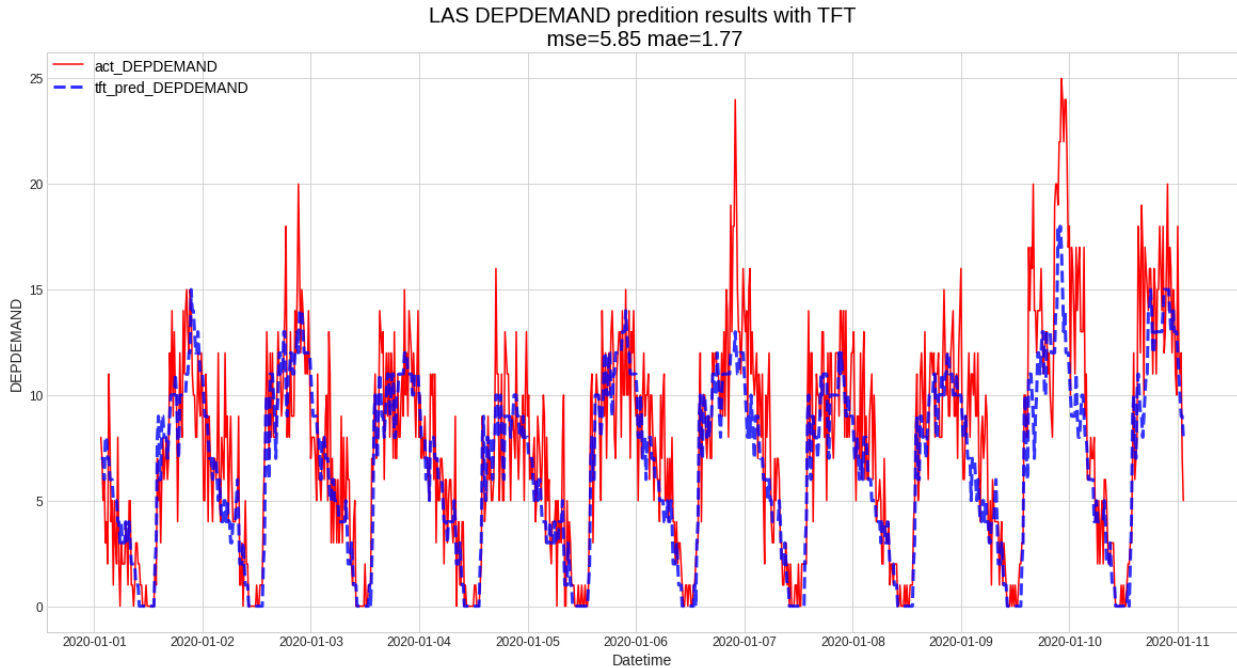

**Fig. 13 LAS quarter-hour departure demand forecasting by TFT model with both ASPM and SWIM data.**

Fig. 14 compares the six models' forecasted quarter-hour departure demand at LAS airport with true demand (red line). According to Fig. 14**Error! Reference source not found.**, only the TFT model combining ASPM with SWIM near-live data can capture the sudden demand surge on 1/10/2020 (highlighted in the orange box).

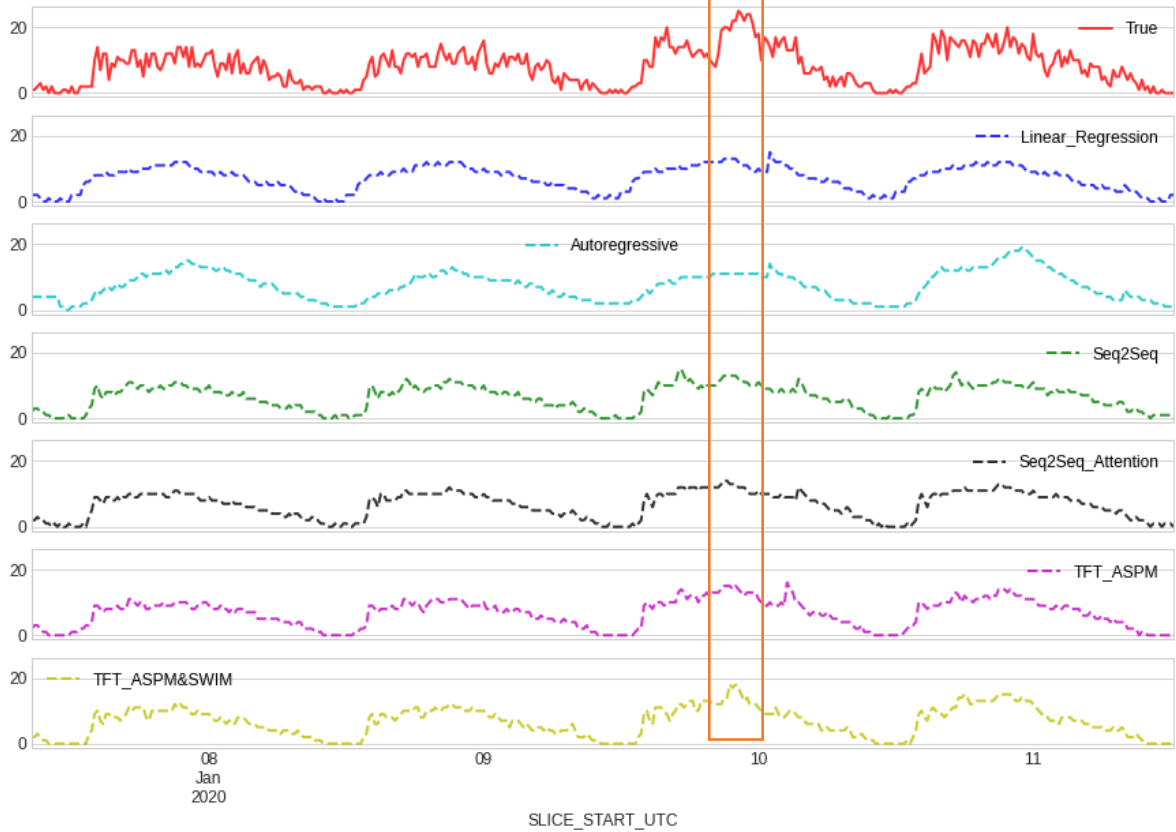

**Fig. 14 LAS models quarter-hour departure demand forecasting results comparison.**

We aggregated the quarter-hour forecasted demand to hourly and daily forecasts. Fig. 15 and Fig. 16 show the comparison of the hourly and daily forecasting results. According to the evaluation metrics, the TFT model with both ASPM and SWIM data is still the closest to true in hourly prediction cases. In comparison, AR is still the least representative of true demand. However, for daily demand forecasting (Fig. 16**Error! Reference source not found.**), after aggregation, the sample size is less than 30 (days), which could be too small to be conclusive on which algorithm performs the best.

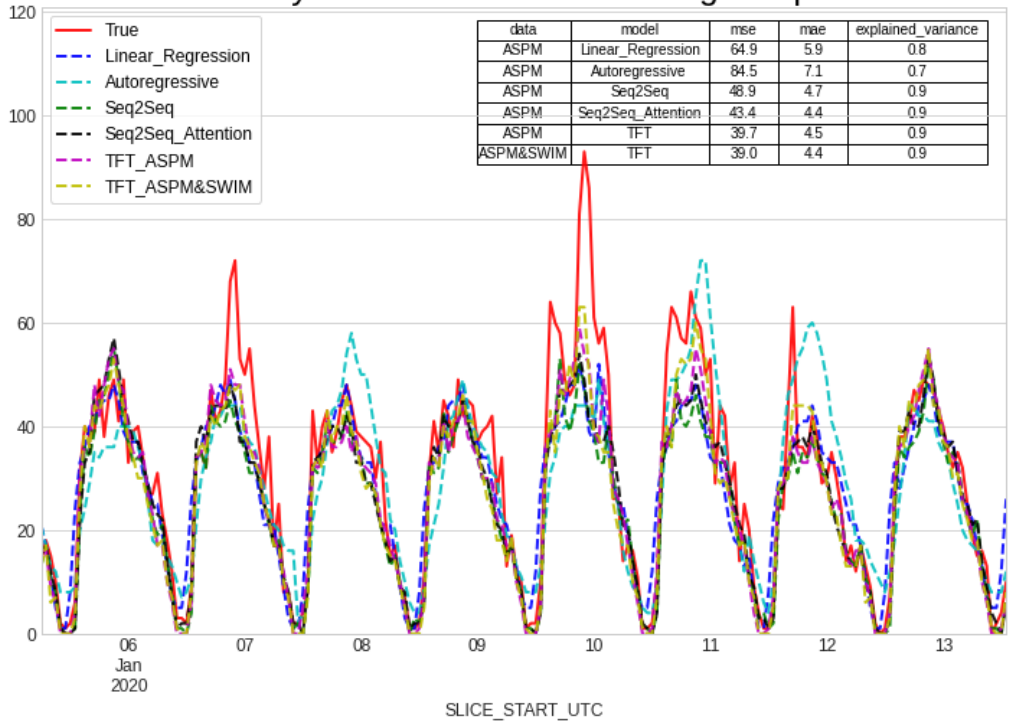

**Fig. 15 LAS hourly departure demand forecasting results comparison.**

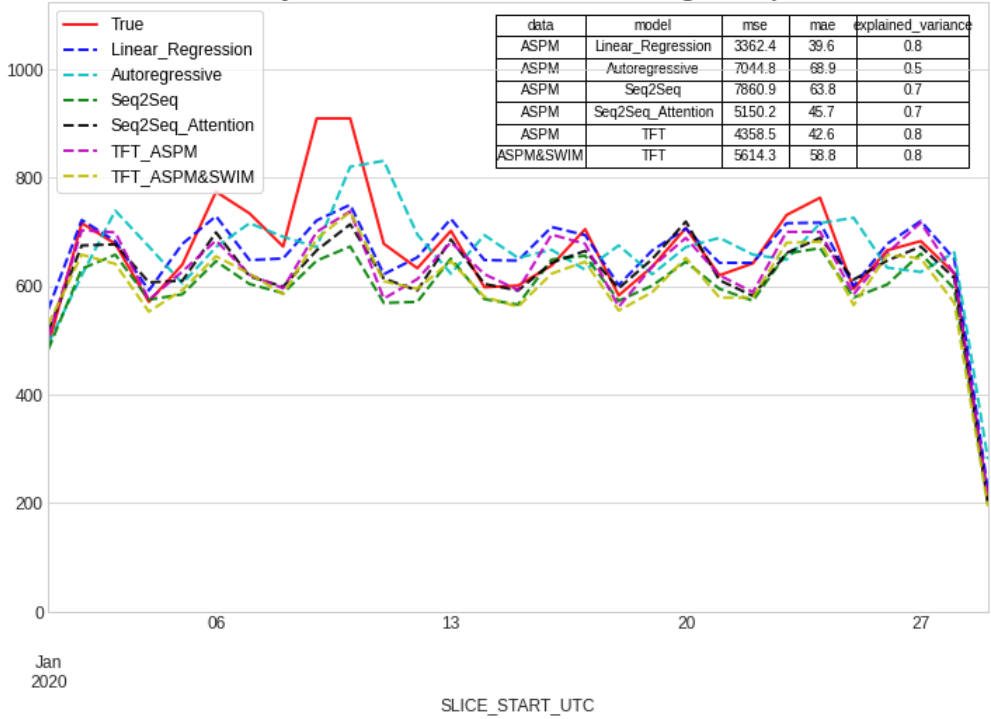

**Fig. 16 LAS daily departure demand forecasting results comparison.**

In short, the LAS case study demonstrates that DL models perform much better than the traditional forecasting techniques, and the TFT model with both ASPM and SWIM data sources performs the best within the different variations we tested.

**2. Model Interpretability**

One nice feature of TFT is that it can examine the importance of each input variable in the prediction. For example, from Fig. 17, important input variables for departure demand prediction can be identified. In this example case, using the TFT model with ASPM and SWIM data as input, the observed departures from the SWIM data and hour input variables are the top 2 factors. Also, the day of the week appears to be less important than the month.

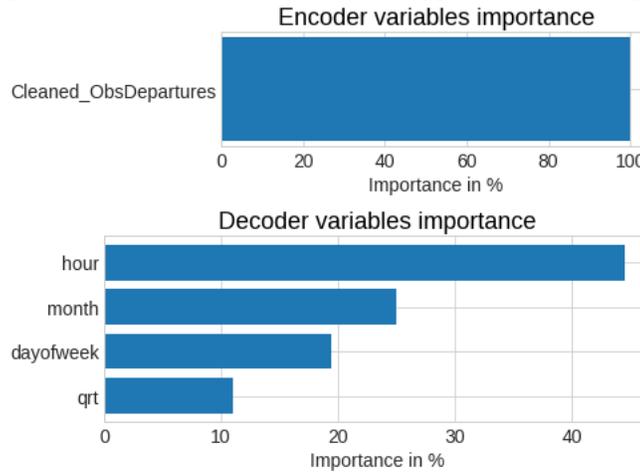

Fig. 17 TFT interpretability with ASPM and SWIM as inputs.

The second interpretability feature is identifying persistent temporal patterns. For example, Fig. 18 shows the attention score relative to time index. A higher attention score means a higher contribution from that time step to the prediction outputs. It should be noted that time index - 1 represents t-1 timestamp's importance for the predictions. As the graph clearly demonstrates, the nearest times play bigger roles, which is as expected.

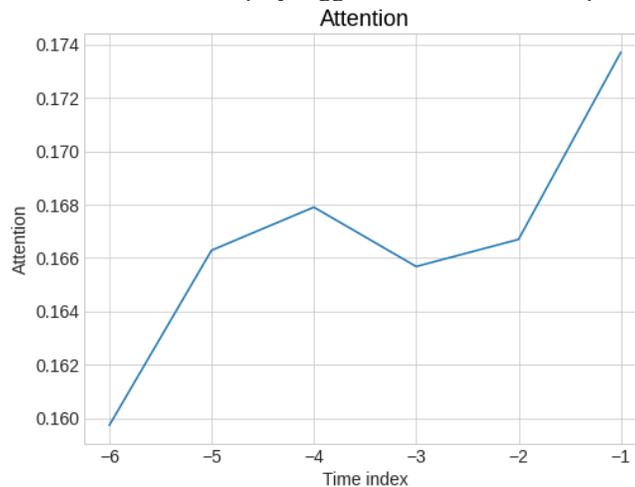

Fig. 18 Attention score by time index.

### C. CLT Departure Demand Forecasting Results

CLT is another airport where Pacer has been implemented. Fig. 19 shows the prediction results for CLT using a TFT model with both ASPM and SWIM as input data sources. From the graph, the TFT model trained for CLT almost perfectly captures the operational banks at CLT airport. With consideration of large banks at CLT, the small mae of 2.59 flights clearly demonstrates the good performance of TFT.

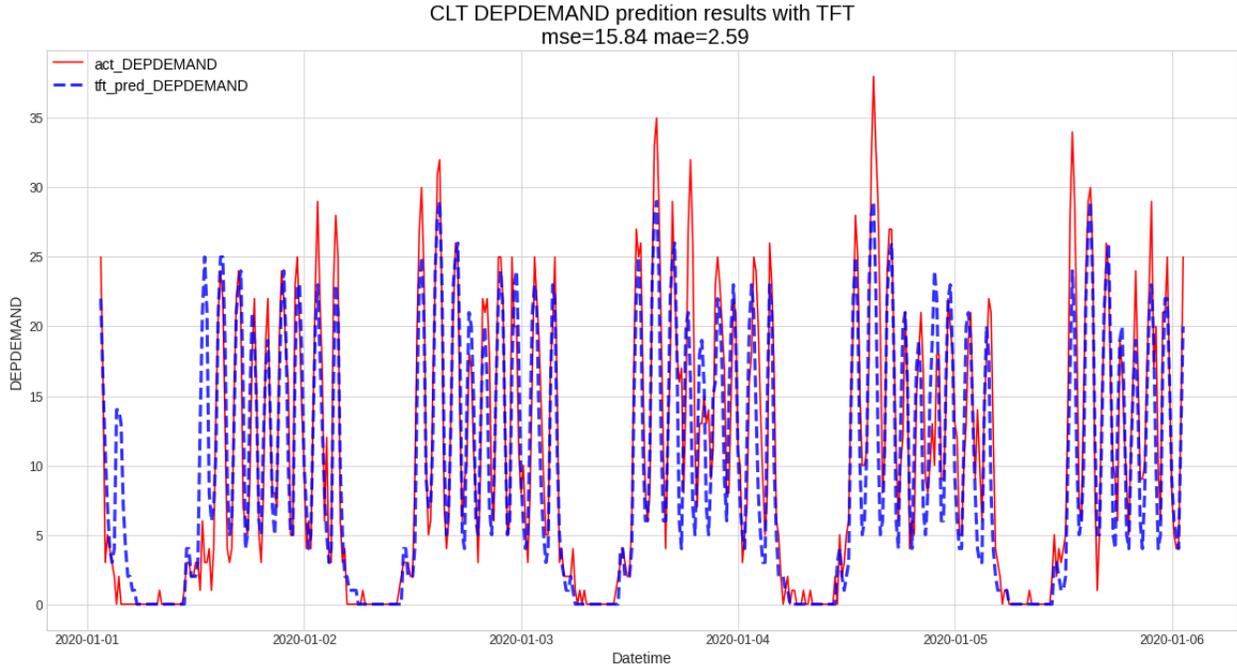

**Fig. 19 CLT quarter-hour departure demand forecasting by TFT model with both ASPM and SWIM data.**

### D. DAL Departure Demand Forecasting Results

We were also interested in improving demand predications at DAL airport. However, processed SWIM data is not currently available for DAL, so we trained a TFT model for DAL with only ASPM data. Fig. 20 shows the prediction results for several testing days at DAL. The figure illustrates that the TFT algorithm accurately captures the departure demand trend throughout the day and does so with small prediction errors of 1.15 flights for mae.

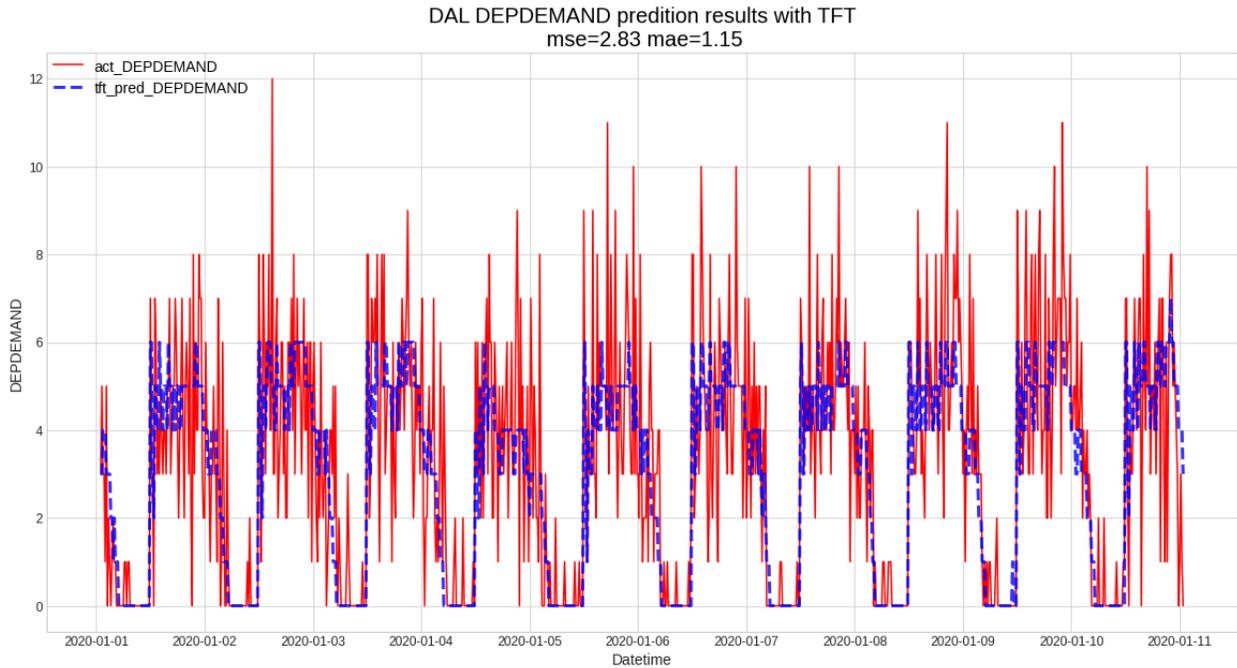

**Fig. 20 DAL quarter-hour departure demand forecasting by TFT model with only ASPM data.**

### E. TEB Departure Demand Forecasting Results

As mentioned above, Teterboro Airport (TEB) has a consistently high proportion of GA traffic. Fig. 21 presents TFT model prediction results when trained and applied to TEB. Similarly, the figure illustrates that the TFT model can capture the demand trend.

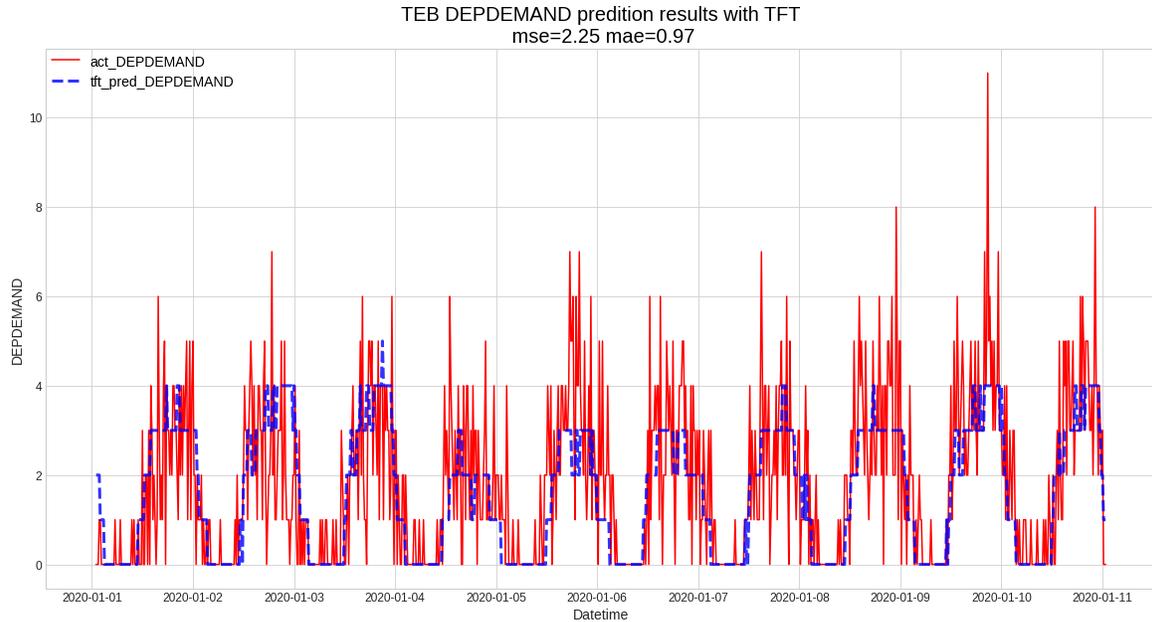

**Fig. 21 TEB quarter-hour departure demand forecasting by TFT model with only ASPM data.**

### F. VNY Departure Demand Forecasting Results

Van Nuys Airport (VNY) also has a high proportion of GA traffic and its variance in demand is very high. Fig. 22 presents the TFT model prediction results when trained and applied to VNY. With a large variance from one quarter-hour to the next and inconsistent patterns to learn at VNY, the prediction results are not as accurate as the above four cases. Actual demand in Fig. 22 can be seen as large up and down swings within each bank of departures. However, the predicted demand generated by the model does not reflect those swings. This would suggest that airports having an irregular operational pattern of many flights followed by very few flights within a short period of time may be difficult for the TFT to model accurately. For these types of airports, further exploration of additional TFT techniques, such as predicting a range at each time step instead of a point estimation, may be needed as that approach may be better suited for fluctuating data patterns.

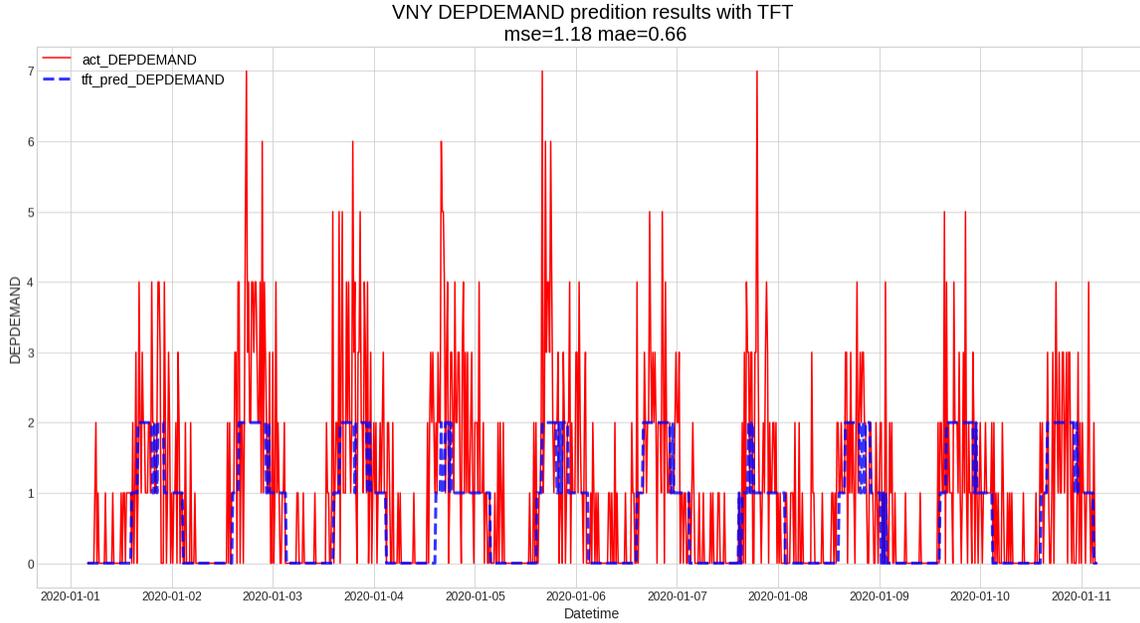

**Fig. 22 VNY quarter-hour departure demand forecasting by TFT model with only ASPM data.**

Through the demonstration of these modeling results, we begin to understand the power and benefits of TFT models. They can perform better than traditional forecasting methods by large margins, and they can result in better predictions of departure demand across diverse airports and with better interpretability.

## VII. Conclusions

This paper applied an attention based neural network model, TFT, to improve flight departure demand forecast accuracy for Pacer, whose objective is to improve surface situation awareness and operational predictability at airports with a sizeable number of GA flight operations that experience departure delays and surface congestion. The research has demonstrated that the combination of good quality input data and advanced forecasting algorithms is a viable path towards obtaining improved prediction performance. Of the five algorithms we tested, TFT predicted departure demand most accurately. TFT models can perform better than the traditional forecasting methods of linear regression and autoregression by large margins (31-53%). In addition, TFT generally produced good predictions of departure demand across the set of diverse airports that our team tested.

Although our current TFT models output one prediction value (50 percentile) at each time step, TFT can output multiple quantile predictions simultaneously. In future work, we can fully utilize this capability to output intervals (e.g., 25th percentiles, median, 75th percentile) across all prediction horizons, which may be useful given the volatility of demand at GA airports.

## Acknowledgments

We thank the following MITRE colleagues: Paul Diffenderfer, Kevin Long, Joey Menzenski, Dr. Ronald Chong, Diane Baumgartner, Caroline Abramson, Dr. Emily Stelzer, and Dr. Panta Lucic for their valuable discussions and insights.